\documentclass[10pt, twocolumn, letterpaper]{article}

\usepackage{cvpr}
\usepackage{times}
\usepackage{epsfig}
\usepackage{graphicx}
\usepackage{amsmath}
\usepackage{amssymb}
\usepackage{float}
\usepackage{subcaption}
\usepackage{booktabs}
\usepackage{stfloats} 
\usepackage[pagebackref=true,breaklinks=true,letterpaper=true,colorlinks,bookmarks=false]{hyperref}
\usepackage{graphicx}
\usepackage{multirow}
\def\BibTeX{{\rm B\kern-.05em{\sc i\kern-.025em b}\kern-.08em
    T\kern-.1667em\lower.7ex\hbox{E}\kern-.125emX}}

\cvprfinalcopy 

\def\cvprPaperID{****} 

\ifcvprfinal\fi

\begin{document}

\title{Knowledge AI: Fine-tuning NLP Models for Facilitating Scientific Knowledge Extraction and Understanding}
\author{Hayden Beadles\dag, Kalyan Sashank Mupparaju\dag, Balaji Muralidharan\dag, Mahmoodreza Marzban\dag\\
{\tt\small beadles@gatech.edu, kmupparaju3@gatech.edu, bmuralidharan3@gatech.edu, mmarzban3@gatech.edu}\\
\dag Georgia Institute of Technology, GA, USA}
\maketitle

\begin{abstract}

This project investigates the efficacy of Large Language Models (LLMs) in understanding and extracting scientific knowledge across specific domains and to create a deep learning framework: Knowledge AI. As a part of this framework, we employ pre-trained models and fine-tune them on datasets in the scientific domain. The models are adapted for four key Natural Language Processing (NLP) tasks: summarization, text generation, question answering, and named entity recognition. Our results indicate that domain-specific fine-tuning significantly enhances model performance in each of these tasks, thereby improving their applicability for scientific contexts. This adaptation enables non-experts to efficiently query and extract information within targeted scientific fields, demonstrating the potential of fine-tuned LLMs as a tool for knowledge discovery in the sciences.
\end{abstract}

\section{Introduction/Background/Motivation}
Large Language Models (LLMs) are increasingly utilized across various Natural Language Processing (NLP) tasks. Our project focuses on developing a deep learning-based framework designed to make scientific text accessible to a general, non-scientific audience. \cite{naveed2024comprehensive} We investigate this by developing an Artificial Intelligence (AI) framework comprised of several NLP tasks. This framework allows individuals to ask questions, identify relationships within the text, and reason about the scientific content without requiring specialized scientific knowledge.\cite{groeneveld2024olmo}
We leverage the capabilities of existing LLMs to perform these NLP tasks effectively on scientific data. We fine-tune these models on specific NLP tasks using relevant scientific data. The four core NLP tasks we focus on this study are: text generation, Question and Answer (Q \& A), summarization and Named Entity Recognition(NER). 

The emergence of LLMs has fundamentally transformed the way information is processed and disseminated. These powerful AI models possess an unparalleled ability to analyze and understand vast amounts of data, revolutionizing both information retrieval and content creation. However, within the realm of scientific research, a critical gap exists – a lack of accessible tools that effectively bridge the communication divide between researchers and the broader public. \cite{boyko2023interdisciplinary} Scientific research often generates complex findings and specialized knowledge that can be challenging for non-experts to grasp. This hinders the dissemination of scientific ideas and discoveries to the general public, limiting their potential impact and hindering broader societal understanding. To address this challenge, there is an urgent need for a unified AI framework specifically designed for scientific tasks. Such a framework would leverage the capabilities of LLMs to enable users to simplify the inference of scientific data, facilitate scientific communication, and increase the accessibility of scientific knowledge. While there are attempts to create open, accessible frameworks, no unified, dedicated framework has been generally accepted. \cite{groeneveld2024olmo}

Government agencies like the Department of Energy (DOE) recognize the potential of AI in facilitating scientific progress, as evidenced by their recent calls for proposals to build tools for interdisciplinary research \cite{DOE_Topics_2024}.   
We discovered only recently that Meta AI also released an LLM, Galactica in 2022\cite{taylor2022galactica} specifically designed for scientific tasks. The motivations for the models were similar in spirit to our goals of knowledge extraction and dissemination. However, due to critical feedback, the Galactica model was taken down shortly after it was unveiled. This highlights the challenges associated with using LLMs for scientific tasks. \cite{techreview2022meta}  Despite these challenges, we believe that with further development, our AI tool can address some of the key issues related to scientific information inference.

The details related to the model and relevant datasets are discussed in the next section along with a brief description of our technical approach related to the four NLP tasks. 

\section{Approach} \label{approach_section}

\subsection{Task Overview}
In the summarization task, advanced models are fine-tuned using an Adaptive Tokenization Strategy to address challenges in scientific text summarization. This approach leverages robust sequence-to-sequence capabilities and the ability to manage long documents, enhancing the accuracy and coherence of the summaries. Further details and comparisons are provided in section \ref{summarization_section}. 

Text generation explores additional techniques such as Low Rank adaptations in training causal language models, and compares and evaluates models that utilize these newer techniques against fine-tuned models in causal language generation. This is discussed in section \ref{text_generation_section}. 

Question answering explores Extractive and Abstractive QA by testing extractive k-shot learning with fine-tuned and generative models, as well as abstractive QA to better understand the model's latent ability to answer short-form questions. Extractive QA results are explored \ref{extract_qa_section} and Abstractive QA \ref{abstract_qa_section}. 

Finally, NER examines how well, fine-tuned models can perform token classification of named entities on scientific datasets of increasing complexity. For NER, we compare and contrast the performances of the models in that area. One important aspect that we discuss in the results section is the influence of class imbalance and its correlation with the dataset size for NER performance. The results are discussed in \ref{ner_section}. 

\subsection{Code / Data} \label{code_section}

Summarization and text generation utilize the scientific papers dataset from the Hugging Face \cite{huggingface2023scientificpapers} to help generate and fine-tune models on scientific data. Summarization examines fine-tuned BART and LED models \cite{huggingface2023bart} \cite{huggingface2023led}. 

Text generation uses distilgpt2 \cite{huggingface2023distilgpt2}. For text generation, we adapted the training framework from Hugging Face's language modeling notebook \cite{huggingface2023languagemodeling} to our particular task. We modified the original code to include LoRA adapters\cite{hu2021lora}.

Question answering fine-tunes two sets of models, BERT and SciBERT are fine-tuned on SQUAD for Extractive \cite{huggingface2023squad}, and PubMedQA for Abstractive \cite{huggingface2023pubmedqa}, on the \emph{question} and \emph{long\_answer} columns, which reflect short form question / answer pairs. For abstractive, SciBERT and BERT are tuned with the EncoderDecoder class in hugging face. \cite{huggingface2023encoderdecoder}, which allows them to perform generative tasks.

NER fine-tunes BERT, SciBERT, and SciDeBERTa against a set of increasing science-focused NER datasets, CoNLL2003, SciERC and GENIA. \cite{huggingface2023conll2003} \cite{luan2018multi} \cite{huggingface2023genia}

Our models use the hugging face set of libraries to build, fine-tune models, extract datasets, and evaluate them on the selected task. The code used to deploy these models and fine-tune them are stored in the Georgia Tech Github organization here: \cite{cs7643team12023} . The code is also included as supplemental material to this paper.

As mentioned, we utilized tools like PEFT and LORA to further explore the refinement of models and also tried multiple approaches in data collection, fine-tuning preparation, and worked through several iterations of work in each task. These challenges we encountered, and worked through, are discussed in more detail in section \ref{experience_challenges_section}. 

\subsection{Metrics}
We employ different metrics to assess the finetuning and model performance on the respective tasks. For text generation, summarization, and Q \& A, we measure the various ROUGE (Recall-Oriented Understudy for Gisting Evaluation) scores. Specifically, we assess the three commonly used ROUGE metrics: ROUGE-1, ROUGE-2, and ROUGE-L. For summarization, we also employ the 'METEOR' metric, which offers a more nuanced evaluation of the summary quality. A human evaluation criterion is used in addition to the ROUGE scores for the Q \& A task to better represent the model's capability to provide scientifically relevant answers. This is a simple score between 0 and 1. Finally, for the NER task, we employ the standard F1, recall, and precision metrics computed based on the total number of predicted entities, the total number of ground truth entities, and the total number of correct entities.
Per-class measures of the F1, recall, and precision metrics were also used to determine the NER class-wise performance.

\section{Experiments and Results}
\subsection{Summarization} \label{summarization_section}
In line with \ref{code_section}, our project centers on deploying and fine-tuning the BART and LED models on a curated scientific article dataset. BART, renowned for its bidirectional encoder and auto-regressive decoder, excels in summary generation, while LED, an extension of the Longformer model, efficiently handles lengthy documents. Fine-tuning offers advantages like task-specific optimization, data efficiency, and high performance. In addressing the challenge of training with limited computational resources on the extensive Arxiv dataset, comprising 1.7 million articles, we sampled 2\% (around 34,000 articles) to manage a representative subset effectively.
We implemented a dynamic tokenization strategy to boost training efficiency, focusing on the top 500 significant tokens from each article within our sampled subset. This streamlined approach reduces computational load and sequence lengths, using token frequency analysis to accelerate training cycles while maintaining focus on relevant content. It effectively addresses challenges like vanishing gradients, enhancing model generalization, and capturing crucial thematic elements in scientific text.
The evaluation of BART and LED models provides valuable insights into their performance across various metrics. Initially, the BART model demonstrates moderate precision, ranging from 44.83\% to 53.14\% for ROUGE-1, indicating its effectiveness in capturing important content from source articles. However, its recall is relatively low, suggesting a potential trade-off between conciseness and comprehensiveness. Fine-tuning BART leads to significant improvements across all metrics, with precision notably increasing to 65.23\% for ROUGE-1, indicating enhanced relevance and accuracy of summaries. On the other hand, the LED model exhibits high recall but lower precision, capturing a broad range of content at the expense of including possibly extraneous information. Fine-tuning LED enhances precision while maintaining comprehensive content capture, albeit with a slight decrease in METEOR score, suggesting a trade-off between precision and linguistic features. Overall, both models show promise after fine-tuning, with improved summarization capabilities suitable for scientific dissemination.
Table~\ref{tab:summary_metrics} summarizes the evaluation metrics for the BART and LED models.

\begin{table}[ht]
\centering
\caption{Summary of Evaluation Metrics for BART and LED Models}
\label{tab:summary_metrics}
\resizebox{\columnwidth}{!}{%
\begin{tabular}{|c|c|c|c|c|}
\hline
\multirow{2}{*}{\textbf{\Large Model}} & \multirow{2}{*}{\textbf{\Large Metric}} & \multirow{2}{*}{\textbf{\Large Precision (\%)}} & \multirow{2}{*}{\textbf{\Large Recall (\%)}} & \multirow{2}{*}{\textbf{\Large METEOR (\%)}} \\
 &  &  &  &  \\
\hline
\multirow{4}{*}{\textbf{\Large BART}} & \textbf{ROUGE-1} & \textbf{44.83 -- 53.14} & \textbf{14.41 -- 18.66} & \textbf{-} \\
 & \textbf{ROUGE-2} & \textbf{8.21 -- 16.65} & \textbf{2.77 -- 5.08} & \textbf{-} \\
 & \textbf{ROUGE-L} & \textbf{29.14 -- 35.55} & \textbf{9.65 -- 11.93} & \textbf{-} \\
 & \textbf{ROUGE-Lsum} & \textbf{36.26 -- 46.02} & \textbf{12.10 -- 15.22} & \textbf{-} \\
 & \textbf{Overall} & \textbf{-} & \textbf{-} & \textbf{11.66} \\
\hline
\multirow{4}{*}{\textbf{\Large BART (Fine-Tuned)} }& \textbf{ROUGE-1} & \textbf{51.33 -- 65.23} & \textbf{24.12 -- 38.57} & \textbf{-} \\
 & \textbf{ROUGE-2} & \textbf{18.91 -- 32.31} & \textbf{9.55 -- 16.14} & \textbf{-} \\
 & \textbf{ROUGE-L} & \textbf{32.94 -- 42.07} & \textbf{15.21 -- 26.75} & \textbf{-} \\
 & \textbf{ROUGE-Lsum} & \textbf{45.28 -- 59.71} & \textbf{21.81 -- 34.18} & \textbf{-} \\
 & \textbf{Overall} & \textbf{-} & \textbf{-} & \textbf{23.31} \\
\hline
\multirow{4}{*}{\textbf{\Large LED}} & \textbf{ROUGE-1} & \textbf{15.43 -- 26.17} & \textbf{56.10 -- 64.77} & \textbf{-} \\
 & \textbf{ROUGE-2} & \textbf{5.73 -- 10.82} & \textbf{20.19 -- 26.63} & \textbf{-} \\
 & \textbf{ROUGE-L} & \textbf{7.97 -- 14.09} & \textbf{28.90 -- 34.10} & \textbf{-} \\
 & \textbf{ROUGE-Lsum} & \textbf{10.85 -- 19.50} & \textbf{40.00 -- 46.04} & \textbf{-} \\
 & \textbf{Overall} & \textbf{-} & \textbf{-} & \textbf{33.34} \\
\hline
\multirow{4}{*}{\textbf{\Large LED (Fine-Tuned)}} & \textbf{ROUGE-1} & \textbf{47.98 -- 60.92} & \textbf{18.98 -- 25.78} & \textbf{-} \\
 & \textbf{ROUGE-2} & \textbf{14.71 -- 25.34} & \textbf{5.72 -- 9.74} & \textbf{-} \\
 & \textbf{ROUGE-L} & \textbf{30.17 -- 40.67} & \textbf{11.95 -- 17.50} & \textbf{-} \\
 & \textbf{ROUGE-Lsum} & \textbf{37.81 -- 51.53} & \textbf{15.01 -- 21.16} & \textbf{-} \\
 & \textbf{Overall} & \textbf{-} & \textbf{-} & \textbf{15.87} \\
\hline
\end{tabular}%
}
\end{table}
In the appendix, Section~\ref{Summarization_Model_Output} showcases summarization outputs from the BART and LED models, using this paper as input. The outputs highlight differences in their performance and content delivery. The fine-tuned BART model produces precise, accessible summaries that effectively encapsulate the goals and methods of leveraging LLMs for NLP tasks, making it ideal for conveying scientific knowledge to non-experts. In contrast, the LED model, while thorough, may include extraneous details that potentially confuse non-specialist readers, illustrating a trade-off between detail and clarity.

\subsection{Text Generation} \label{text_generation_section}
In this part of our project, we focused on enhancing the text generation capabilities of a transformer-based language model, specifically using the distilgpt2 model for its computational efficiency. From the ArXIV dataset we sampled 10,000 training papers, 1,000 validation, and 1,000 test papers, processed into 256-token segments for causal language modeling. We employed two fine-tuning methods: full fine-tuning and Parameter-Efficient Fine-Tuning (PEFT) with LoRA adapters(Low-Rank Adaptation of LLMs)\cite{hu2021lora}. Full fine-tuning was expected to deeply integrate the scientific terminology and style into the model due to the comprehensive update of all parameters. In contrast, PEFT aimed to offer significant performance enhancements with less computational overhead by updating only specific low-rank matrices within the model's architecture. 

We set up our experiments to generate text continuations from scientific paper prompts. Success was measured using ROUGE scores to quantitatively assess the overlap between the generated text and ground truth texts from scientific papers. As seen in Table~\ref{tab:comparative_performance_generation}, the full fine-tuning method resulted in higher ROUGE scores, indicating better performance in generating accurate and relevant text. PEFT also improved performance over the baseline but to a lesser extent. Despite the lower improvement, PEFT's efficiency in resource use was a notable advantage.

Qualitatively, we evaluated the coherence and contextual relevance of the generated texts. Examples of model-generated text are in the appendix section \ref{Generation_Output} for comparison. Our results show that full fine-tuning yielded the most coherent and contextually appropriate text. Although PEFT was slightly less effective, it still marked an improvement over the baseline and used significantly fewer resources. Both methods successfully enhanced the model's capability to generate scientific text, confirming our hypothesis about the benefits of fine-tuning. 

\begin{table*}[ht]
\centering
\caption{Comparative performance, training time, and parameters trained for baseline, full fine-tuning, and PEFT with LoRA adapters on text generation. The table highlights the efficiency of PEFT, which approaches full fine-tuning performance with significantly fewer parameters and reduced training time.}
\label{tab:comparative_performance_generation}
\scalebox{0.75}{ 

\begin{tabular}{lccccc}
\hline
\multicolumn{1}{c}{\textbf{Model}} & \textbf{ROUGE-1 F1} & \textbf{ROUGE-2 F1} & \textbf{ROUGE-L F1} & \textbf{Training Time (hh:mm:ss)} & \textbf{Parameters Trained} \\
\hline
\textbf{Baseline} & 0.222319 & 0.025179 & 0.135648 & - & - \\
\textbf{Full Fine Tuning} & 0.268315 & 0.036105 & 0.157470 & 7:05:03 & 82M \\
\textbf{PEFT Fine Tuning} & 0.252146 & 0.031211 & 0.151372 & 5:09:49 & 294,912 \\
\hline
\end{tabular}
}
\end{table*}

\begin{table*}
    \centering
    \caption{K Shot Extractive QA Model Performance Comparison (Averaged over Easy, Medium, Hard questions)}
    \label{tab:extractive_finetune_comparison}
    \scalebox{0.75}{ 
        \begin{tabular}{lcccc}
            \hline
            Method & Scibert FineTune & Bert FineTune & Google T5 & T5-Flan \\
            \hline
            \textbf{Rouge Score - Few Shot} & .74833 & .689 & .0267 & .467  \\
            \textbf{Correctness - Few Shot} & 1 & .8333 &  .3 & .95 \\
            \textbf{Rouge Score - One Shot} & 0.7622 & 0.6106 & 0.0286 & 0.244 \\
            \textbf{Correctness - One Shot} & 1 & 0.667 & 0.05 & 0.833333333 \\
            \textbf{Rouge Score - Zero Shot} & 0.689 & 0.222333333 & 0 & 0.466666667 \\
            \textbf{Correctness - Zero Shot} & 0.833333333 & 0.416666667 & 0.083333333 & 1 \\
            \hline
            Summary (.6 * Human + .4 * Rouge) & .84993 & .5695  & .0940  &.7136  \\
            \hline
        \end{tabular}
    }
\end{table*}

\subsection{Extractive QA - K-Shot Results} \label{extract_qa_section}
\begin{figure*}[t] 
  \centering
    \centering
    \includegraphics[width=.6\linewidth]{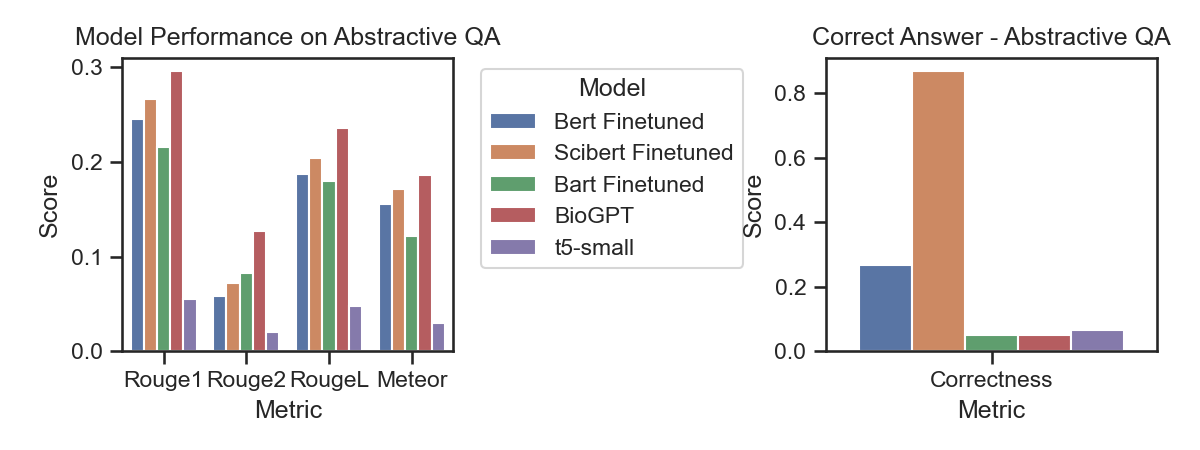} 
    \caption{\small In the left image, we show test set ROUGE and METEOR performance on the PubMedQA dataset. Of these, BioGPT appears to do best, but it is misleading. We took a sample of the test set and compared answers in the image on the right. BioGPT and BART repeat the question in the answer, and only SciBERT appears strongest and responds to the question. It scored well in this particular sample of 10 questions. Please go here  \ref{abstractive_qa_output} to see output in model question / answers. }
    \label{fig:abstractive_qa_results}
\end{figure*}
As shown in table \ref{tab:extractive_finetune_comparison}, as the number of K-shot examples provided to the models gets smaller, the importance of strong pre-training in a relevant domain via the base model becomes evident. Of all these, SciBERT performs the strongest throughout K-shot scenarios, while BERT drops off in performance. This indicates that SciBERT has more of an understanding of the underlying subject, than simply gaming out the Question / Answer strategy for the task. \cite{chada2021fewshotqa}

Generative models show a similar story. (see table \ref{tab:extractive_finetune_comparison}). Of all the models, T5 does the weakest. It is clear that the base model is not trained in a way where it can deduce the answer or handle the task. FLAN-T5 is strong, able to maintain strong scores in both ROUGE and correctness. The pre-training done on the FLAN-T5 model has enabled it to tackle the Extractive QA task without fine-tuning. \cite{luo2022choose}

Overall, when formed on scientific data, models pre-trained on scientific data, and then fine-tuned on the QA task, seem to hold the most consistency and relevance in extracting the answer from the provided context. Please visit Appendix section \ref{extractive_qa_explanation} for more background on the task.

\subsection{Abstractive QA - Generative Results}\label{abstract_qa_section}

As discussed in \ref{code_section}, we fine-tuned two Encoder / Decoder versions of Bert and SciBERT using huggingface and trained them on PubMedQA. BioGPT and BART are community models trained on the question and \emph{final\_answer}, which contain either yes or no. Can these models adapt to short QA pairs? T5 is included as a baseline. 

The results of testing these models on PubMedQA can be seen in Fig.\ref{fig:abstractive_qa_results}. Two images are shown. The image on the left contains the metric performance of various models, and the one on the right shows a sample performance of the models where we manually reviewed each provided answer for correctness. To view comparison in abstractive model output, please go here \ref{abstractive_qa_output}. 

The results suggest that, while BART and BioGPT appear to perform well on ROUGE and METEOR \ref{fig:abstractive_qa_results}, they cheat in their answers; repeating the question in the answer. This most likely has to do with the shorter max length parameter in these models' training.

Fine-tuned SciBERT and BERT models show a better understanding of medical domain questions, with SciBERT, in particular, demonstrating strong performance and evidence of transfer learning from its extensive pre-training on scientific papers. In the referenced figure, SciBERT does very well on the sample of questions we evaluated (10 questions from the test set) \ref{fig:abstractive_qa_results}. 

Additionally, it shows the flexibility of a model like SciBERT in pre-training and fine-tuning. Using the model in a generative form (via the EncoderDecoder class in hugging face), still retains the models ability to form strong answers on difficult, nuanced topics.

\subsection{Named Entity Recognition} \label{ner_section}
Named Entity Recognition (NER) is an NLP task of classifying words present in an input text as belonging to different labeled entities. Principally, NER is a token classification task that associates various text tokens to different classes. NER could be a powerful technique to parse through scientific text and extract the various associations related to specific scientific concepts. Extracting the specific salient information from a list of papers would be very useful to filter and identify only the relevant papers for further study.
Our goal for this project task is to perform NER of scientific text using pre-trained LLMs. We finetune various pre-trained LLMs using scientific datasets and assess their performance.  This approach allows us to investigate the effectiveness of different LLM architectures and their ability to adapt to the specific domain of scientific language. We perform fine-tuning for NER tasks using three pre-trained LLMs: BERT, SciBERT and SciDeBERTa. Since all the LLMs are 'Encoder' only architecture, for NER, we use the same baseline architecture with an additional output layer on top of the BERT/SciBERT/SciDeBERTa models. The token classification head applies a linear layer followed by a softmax activation function. The number of output units in the final layer corresponds to the number of token labels present in the NER task. 

As reported earlier, three datasets are used for NER: ConLL2003, SciERC, and GENIA. The only hyper-parameters for the finetuning are the learning rates used for optimization, batch size, and the number of epochs. For the SciERC dataset, all the finetuning cases were run with a learning rate of 2e-5, for 10 epochs. For the ConLL2003 and GENIA datasets, the fine-tuning was performed for 5 epochs. A batch size of 32 was used for the SciERC dataset and 16 was used for the other two datasets. The F1, precision and recall scores based on the validation datasets for BERT, SciBERT and SciDeBERTa are reported in Table. \ref{tab:ner_performance}. 

The observations from the finetuning results can be summarized as follows:
For the general dataset, BERT outperforms SciBERT, as expected due to BERT's pre-training on a large general corpus. SciDeBERTa performs as well as BERT, likely due to its larger number of parameters and pre-training with scientific data. On the SciERC dataset containing scientific text, SciBERT performs better than BERT, highlighting the importance of domain-specific pre-training for specialized datasets, especially when the dataset size is small as for the case that used a reduced dataset size (half the original size). SciDeBERTa achieves the best results overall. For the GENIA dataset, both BERT and SciBERT perform well, underscoring the importance of dataset size – a sufficiently large dataset can compensate for BERT's lack of initial scientific knowledge. The observations highlight the trade-offs between general and domain-specific pre-training, as well as the role of dataset size.

\begin{figure}[H]
    \centering
    \includegraphics[width=0.7\linewidth]{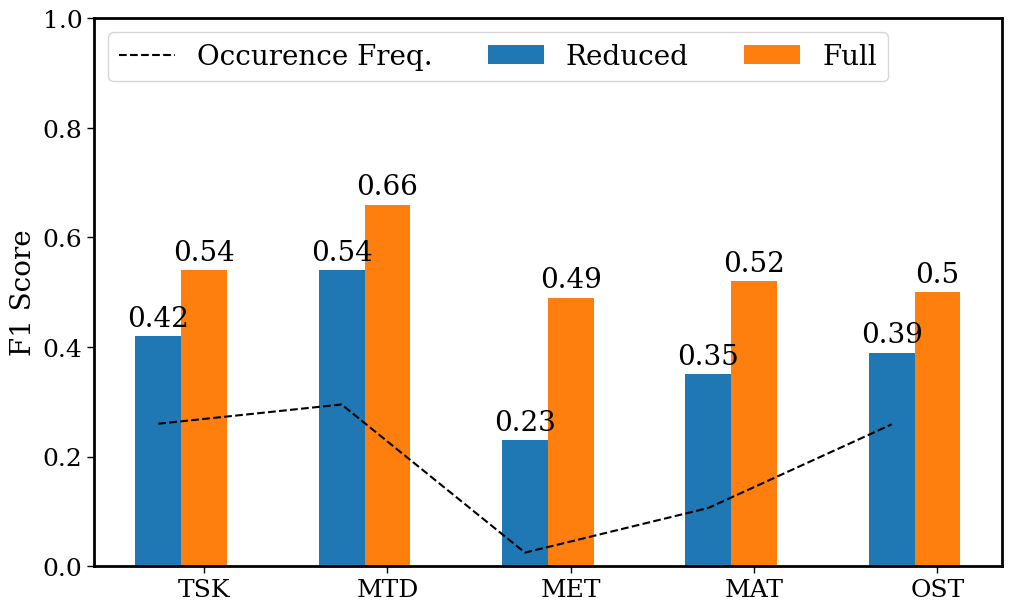}
    \caption{Comparison of F1 score for NER task finetuned with a full and reduced SciERC dataset with BERT model.}
    \label{fig:classimbalance}
\end{figure}
 Furthermore, the impact of class imbalance, i.e., differences in the frequency of occurrence of the named entity labels in the dataset, on NER performance is evaluated and illustrated in Fig. \ref{fig:classimbalance}. We observe that while the F1 scores strongly correlate with the normalized occurrence score for the reduced dataset, this correlation significantly weakens as the dataset size increases.

\begin{table*}
\centering
\caption{NER performance of BERT, SciBERT, SciDeBERTa on CoNLL2003, SciERC, and GENIA datasets on validation set.}
\label{tab:ner_performance}
\scalebox{0.75}{
\begin{tabular}{lcccccccccc}
\hline
\multicolumn{2}{c}{} & \multicolumn{2}{c}{\textbf{BERT}} & \multicolumn{3}{c}{  \textbf{SciBERT}} & \multicolumn{4}{c} {\textbf{SciDeBERTa}} \\
\cmidrule{1-3} \cmidrule{3-5} \cmidrule{6-11}
Type of dataset & Task & F1 & Precision & Recall & F1 & Precision  & Recall & F1 & Precision  & Recall \\ \hline
CoNLL2003 & General  & 93.6 & 92.6  & 94.5 & 84.3 & 83.0  & 86.0 & 93.3 & 92.6 & 94.1\\
~(14 K) & &  &  &  &  &  \\
SciERC & CS  &  59.2 & 55.8 & 63.0  & 60.6  & 58.4 & 63.1 & 62.0 & 59.5 & 64.4\\
~(1.9 K) &  &  &  &  &  &  \\
SciERC-reduced & CS  &  40.8 & 36.8 & 45.7  & 50.8  & 48.3 & 53.4 & 50.2 & 46.8 & 54.1\\
~(1 K) &  &  &  &  &  &  \\
GENIA & Bio & 70.0 & 65.0 &  76.5 & 72.0  & 67.0 & 78.0 & 76.0 & 71.5 & 81.0 \\
~(18.5 K) &  &  &  &  &  &   \\ \hline
\end{tabular}}
\end{table*}

\section{Experience/Challenges} \label{experience_challenges_section}
In this section, we discuss the various observations, our experiences related to fine-tuning large language models, and the associated challenges. Our findings are grouped into categories related to training, data size, and computational resource requirements:
\subsection{Optimizing Training Efficiency on Large Scientific Datasets}
In conjunction with the Adaptive Tokenization Strategy discussed in Section \ref{summarization_section}, we utilized parallel processing during the preprocessing stage. The tokenizer efficiently processed articles and summaries by harnessing the power of up to 16 CPU cores for parallel execution. This method notably slashed preprocessing times, facilitating quicker training and more proficient management of the extensive dataset.

\subsection{Challenges with BART in Long Document Summarization}
In our initial approach to document summarization, we anticipated that BART would be effective in producing high-quality summaries. However, we encountered significant limitations related to BART's maximum input token constraint, which hindered its ability to process longer documents effectively. This challenge compelled us to explore alternative models designed to handle extensive texts. Our research led us to Longformer\cite{beltagy2020longformer} and LED (Longformer Encoder-Decoder), which are specifically tailored for processing longer documents by extending the maximum input length far beyond what BART can manage. The implementation of these models was not straightforward and required adjustments in our preprocessing and model configuration to effectively utilize their extended capabilities. Although this shift was not part of our original plan, it ultimately provided a viable solution to the token limitation issue, illustrating the importance of adaptability in managing large-scale language model applications.

\subsection{Efficiency and Practicality of LoRA in Model Adaptation}
LoRA (Low-Rank Adaptation) introduces low-rank matrices to the self-attention mechanisms of distilGPT-2, altering only 0.36\% of the parameters \cite{hu2021lora}. Initially, we anticipated significant reductions in training time due to fewer trainable parameters. However, the adapted model still required about 70\% of the time needed for full fine-tuning. This lesser-than-expected efficiency can be attributed to the computational complexity of self-attention layers, challenges in integrating LoRA with the existing model structure, and the deep network backpropagation requirements. Moreover, hardware and data processing efficiencies critically influenced training times. Despite these challenges, LoRA proved substantially efficient in parameter utilization and adaptability with constrained resources. The complexities of transformer architectures and the demands of modifying attention mechanisms, however, moderated the time savings. Future efforts could enhance efficiency by optimizing data handling, computational pathways in adapted layers, and improving hardware utilization, aiming to maximize the gains from LoRA adaptations \cite{hu2021lora}.

\section{Conclusions}
We have presented in this study Knowledge-AI, a deep learning framework designed to extract and comprehend content from scientific texts, making it more accessible to a wider audience. We leverage the power of large language models by fine-tuning them for specialized NLP tasks tailored to the scientific domain.  Focusing on four critical NLP tasks – text generation, summarization, question-and-answering, and named entity recognition – we demonstrate the effectiveness of fine-tuning LLMs like BART, BERT, and SciBERT for scientific text analysis. We have summarized some of the key findings related to each of the four NLP tasks:
\begin{itemize}
\item \textit{Summarization}: When summarizing scientific papers, fine-tuning BART and LED models is highly effective. These models excel at capturing fine details, aided by crucial tokenization for training with limited resources. For clear, concise summaries suited to general audiences, opt for the fine-tuned BART model, while the LED model handles more comprehensive content. Our study emphasizes the importance of selecting the right model for effective scientific communication.
\item \textit{Text Generation}: Our experiments demonstrate that full fine-tuning outperforms Parameter-Efficient Fine-Tuning (PEFT) with LoRA adapters in generating text from scientific domains, as shown by higher ROUGE scores. Although PEFT offers substantial computational savings, reflecting a reduced number of trained parameters, it still effectively enhances model performance over the baseline. This highlights PEFT as a viable alternative for scenarios where resource efficiency is prioritized, suggesting potential areas for future optimization to bridge the performance gap with full fine-tuning.
\item \textit{Question / Answering}: Domain pre-training makes a significant impact in K-Shot scenarios for Extractive QA, allowing fine-tuned models like SciBERT strength as representative examples decrease. Additionally, FLAN-T5 shows that generative models can perform well. In Abstractive QA, SciBERT Full Transformer was able to directly answer the short questions, reflecting the interior strength of its training in generation and question answering. Inconsistent behavior was found in the other models
\item \textit{NER}: Domain-specific pre-training provides a significant advantage in the case when the data size is low. However, as the dataset size increases, even models pre-trained with general text, such as BERT, start to perform increasingly well. Similarly, class imbalance issues, which are accentuated in the case of low dataset size, diminish with an increase in data size. 
\end{itemize}

Overall, this study lays a strong foundation for the continued integration of NLP advancements into scientific knowledge dissemination.

\section{Work Division}
We list the division of work between the team in table \ref{tab:contributions}. 
\begin{table*}[ht]
\begin{center}
\begin{tabular}{lcp{8cm}}
\hline
\textbf{Student Name} & \textbf{Contributed Aspects} & \textbf{Details} \\
\hline
Hayden Beadles & Question Answering & Fine-tuned models for Extractive and Abstractive QA using SciBERT, BERT, T5, and FLAN-T5. Evaluated in K-Shot scenarios and implemented LORA. Developed scripts for visualization and validation. \\
Kalyan Sashank Mupparaju & Text Generation, LoRA Fine-Tuning & Focused on full and PEFT using LoRA adapters with the distillgpt2 model. Optimized the model for text generation tasks and adapted it for generating scientific texts from the arXiv dataset. Developed scripts to evaluate the performance of each model. \\
Balaji Muralidharan & NER Tasks, Class Imbalance Research & Developed a streamlined workflow for fine-tuning models on diverse datasets. Created procedures for pre-processing, and set up an inference engine for NER on scientific texts. Established post-processing scripts for result interpretability. \\
Mahmoodreza Marzban & Text Summarization& Fine-tuned BART and LED models for text summarization on Arxiv data. Utilized strategic sampling, dynamic tokenization, and multi-core processing to optimize training and efficacy. Conducted evaluations to assess model performance and task suitability. \\
\hline
\end{tabular}
\end{center}
\caption{Contributions of team members to the project.}
\label{tab:contributions}
\end{table*}


{
\bibliographystyle{ieee_fullname}
\bibliography{egpaper_for_review}
}

\appendix
\section{Summarization: Model Output 
Example}\label{Summarization_Model_Output}
\paragraph{BART Model Summary}
\textit{``Large Language Models (LLMs) are increasingly being used to understand and extract scientificknowledge across specific domains. The emergence of LLMs has fundamentally transformed the way information is processed and disseminated. However, there is a lack of accessible tools that effectively bridge the communication divide between researchers and the broader public.''}

\paragraph{Fine-Tuned BART Model Summary}
\textit{we present a framework for extracting knowledge from scientific text that is accessible to a general, non-scientific audience without requiring specialized scientific knowledge. the framework is comprised of several Natural Language Processing (NLP) tasks that can be performed by large language models (LLMs) trained on specific datasets in the scientific domain and fine-tuned for four key NLP tasks: summarization, text generation, question answering, and named entity}
\paragraph{LED Model Summary}
\textit{``Abstract:This project investigates the efficacy of Large LanguageFilenameModels (LLMs) in understanding and extracting scientific weapknowledge across specific domains. We employed pre-processor-trained models and fine-tuned them on datasets in the sci- warrentific domain. The models were adapted for four key suburNatural Language Processing (NLP) tasks: summariza- warrtion, text generation, question answering, and named entity-recognition. Our results indicate that domain-specific fine-tune and finetuning significantly enhances model performance in each ofthese tasks, thereby improving their applicability for scien- warrtific contexts. This adaptation enables non-experts to effi- warrciently query and extract information within targeted scien - warrtifific fields, demonstrating the potential of fine- tuned LLMs to transform the way information is processed and disseminated intoa tool for knowledge discovery in the sciences. subur1. Introduction/Background/Motivation:Large Language Models (LLM) are increasingly uti- suburlized across various Natural Language Processing(NLP)-tasks. Our project focuses on creating a deep learning- and cognitive-based framework for extracting knowledge from scientific weaptext in a way that is accessible to a general, non-scientific, and/or non-audience. We investigate this by developing an Artificial suburIntelligence (AI) framework comprised of several Natural weapLanguage Processing (LLP) task. This framework allows non-individuals to ask questions, identify relationships within scientific weapthe text, and reason about the scientific content without re- warrquiring specialized scientific knowledge. To perform these weapNLP tasks effectively on scientific data, we leverage the ca- warrpabilities of existing LLMs. We fine-Tune these models on datasets that are not subject to the same constraints as the domain-Specific NLP tasks using relevant scientific data. The four suburcore NLP task we focus on this study are: text generation.Question and Answer (Q \& A), summarization and NamedEntity Recognition(NER). suburThe emergence of LLMs possess an unparalleled ability to an- warrcatalyze and understand vast amounts of data, revolutionizing the way people consume and understand the scientific information. To do these weapwithin the realm of scientific research, a critical gap ex- warrists – a lack of accessible tools that effectively bridge the misunderpabilities between researchers and the broader public. Scientific research often generates complex findings and data that can be challenging for non-scientists to''}
\paragraph{Fine-Tuned LED Model Summary}
\textit{`` we investigate the efficacy of Large Language Models (LLMs) in understanding and extracting scientific information from scien-tific data.  we employ pre-trained models and fine-tuned them on datasets in the sci-fi domain to extract relevant scientific information from scien-tific data.  we''}
\section{Extractive QA - K Shot Model Comparison} \label{extractive_qa_explanation}

We test Extractive QA K-Shot tasks by providing a sequence of gradually more challenging question / answer pairs with decreasing K-shot examples (Starting with few-shot, then one-shot, then zero-shot). The goal here is to gauge how well a model, fine-tuned on SQUAD, and with its underlying pre-training, performs as the complexity of the question increases and the examples provided decrease. In this way, we would hope to see the relationship between a model's pre-training "shine through" the fine-tuning and strengthen the models abilities when it has no examples to rely on. 

We note that we fine-tuned two models, BERT and SciBERT on SQUAD, and used two off-the shelf models that are used for generative tasks (T5, FLAN-T5). 

See table \ref{tab:extractive_finetune_comparison} for a reference of K-shot Extractive QA performance across all models. In table \ref{tab:medium_difficulty_qa}, we give an example of some responses for Few-shot learning with a CyberSecurity domain question (note that a context is provided along with the question but is not included due to length)

\begin{table*}[ht]
\centering
\caption{Medium Difficulty Extractive QA  (Few Shot)- What types of cyber-attacks can Generative AI be used to create? Answer - deepfakes or realistic phishing content}
\label{tab:medium_difficulty_qa}
\begin{tabular}{lllll}
\hline
Task Type   & Model                      & Answer                                    & Rouge Score & Human Review \\
\hline
Extractive  & Scibert - Finetuned QA     & deepfakes or realistic phishing content   & 1         & 1          \\
Extractive  & Bert - Finetuned QA        & such as deepfakes or realistic phishing content               & 0.66         & 1           \\
Abstractive & google/t5-small-ssm-nq     & attack                                    & 0.0         & .15           \\
Abstractive & google/flan-t5-small       & deepfakes or realistic phishing content  & 1         & 1          \\
\hline
\multicolumn{5}{p{16cm}}{\textit{Comments: Bert did just as well as Scibert, surprising! Additionally, the flan-t5-small model got the question correct without any fine-tuning, showing the advantages of the flan t5 model over the base t5 model.}}
\end{tabular}
\end{table*}

What we see from the results is that SciBERT maintains the highest score across correctness (human review, actual answer) and ROUGE as the number of provided examples to the model decrease. This makes it more resilient to changes in the number of representative examples of the task its given. In contrast, BERT's performance decreases. 

As mentioned in the results section, generative models are interesting. T5 performs the poorest. It does not understand the background, and it was not pre-trained on any QA task or scientific dataset. This indicates the danger of simple taking these models for granted as is. It is important to understand each model's training and underlying contributions, making openness critical. 

FLAN-T5 performs very well, built as a model with training on QA. It is not trained on Science data directly, but seems to grasp the task and can still handle the relevant questions.

\section{Text Generation: Model Output Example}
\label{Generation_Output}
This section evaluates the performance of the base, fully fine-tuned, and PEFT(LORA adapter) fine-tuned models in generating text continuations from a scientific prompt related to quasinormal modes and black holes. The prompt and each model's continuation are analyzed to assess their relevance, coherence, and scientific accuracy.

\paragraph{Prompt}
\begin{quote}
    ``the quasinormal modes (qnms) of spacetimes containing black holes have been studied since the pioneering work of Vishveshwara @xcite, who first observed quasinormal ringing of a Schwarzschild spacetime in''
\end{quote}

\paragraph{Base Model Continuation}
\begin{quote}
    ``1905. With the discovery of a quasinormal ring, one could not only speculate on the nature of one's quantum waveforms and the extent of their unique nature—and indeed what is their nature's physical properties.\newline
    \newline
    The quasinormal theory holds an important empirical validity for quantum theory, since it has many characteristics (e.g., quantum law, quantum field theory, or quantum gravity), but not all the ones which define quantum theory. For example, quantum mechanics claims''
\end{quote}
\textbf{Analysis:} The base model introduces irrelevant historical context (``1905'') and diverges significantly from the scientific topic, focusing instead on a broad, somewhat incoherent discussion of quantum theories rather than specifically on quasinormal modes in black holes.

\paragraph{Full Fine Tuned Model Continuation}
\begin{quote}
    ``the presence of a black hole.They discussed the possibility that the frequency of the vibrations in the qnyi modes would lead to a significant modulation of the number of qnms in the presence of the black holes. In this study it was shown that these qnms may play a role only in a small number of quasars. it was confirmed recently that there are many qnms that can not decay in order to reach @xmath39 in the first qne''
\end{quote}
\textbf{Analysis:} This continuation maintains better relevance to the original topic of quasinormal modes and black holes. The use of specific terms like ``vibrations,'' ``qnyi modes,'' and ``qnms'' indicates improved domain-specific knowledge, though the coherence and precision of the continuation could be enhanced.

\paragraph{LORA Adapter Fine Tuned Model Continuation}
\begin{quote}
    ``a zwab - pi particle., @xcite has shown that quasinormal ringing is not completely uniform when the quantum properties are weak and can be treated as @xmath102 in the following cases:.[ fig : wave - vibrational-m ]   because of the quasinormal nature of the vibrational ring, it appears that in the case of the quantum property of @xmath104, the resonant quantum properties of the vibrational ring can be ignored''
\end{quote}
\textbf{Analysis:} This output includes appropriate scientific terminology (``quasinormal ringing,'' ``vibrational ring'') but struggles with coherence and clarity. The text is disrupted by unclear references and lacks specific relevance to the discussion about black holes and quasinormal modes.

\section{Abstractive QA Output} \label{abstractive_qa_output}

In Abstractive QA, we want to test an underlying model's pre-training by only asking the model a question, and expecting a short answer in response. The goal is to see how well the model can extract the answer via its own pre-training, and if it can understand "how" to answer the question. It is a test of a model's ability to compress information in a way that addresses the question. 

We trained BERT an SciBERT as Full Transformers, using the EncoderDecoder class in HuggingFace and then fine-tuning them on PubMedQA (as explained in the Approach section). We then evaluate these models along with some community contributed ones also trained on PubMedQA, but on different columns. The idea is to see the distinction between using an "off-the-shelf" model without understanding the background on the model's training and seeing how it compares to our own. 

Similar to the Text Generative Section, we include some example questions and responses to compare and contrast the Abstractive QA models results when evaluating them for correctness. 

\paragraph{Prompt - Zero Shot}
\begin{quote}
    ``Question: Does macrolide Resistance in Treponema pallidum correlate With 23S rDNA Mutations in Recently Isolated Clinical Strains?. Answer: ''
\end{quote}
\paragraph{BioGPT Answer}
\begin{quote}
    ``Question: Does macrolide Resistance in Treponema pallidum correlate With 23S rDNA Mutations in Recently Isolated Clinical Strains?. Answer: The''
\end{quote}
\textbf{Analysis:} Note that here, BioGPT just repeats the question in its output, it adds only the token at the end, This suggests that it is limited by the training induced on it to only produce very short tokens of a very small length. This limitation essentially prevents the model from answering anything beyond yes / no questions
\paragraph{BART Answer}
\begin{quote}
    ``key clinical message this is the first case of a patient who was diagnosed with''
\end{quote}
\textbf{Analysis:} BART does not appear to be answering the question, just generating tokens to extend the sentence. It also ends up scoring poorly, but it appears to do better than BioGPT. 
\paragraph{BERT Answer}
\begin{quote}
    ``our results suggest that the presence of macrolides in the genome of the zebrafish genome is associated with a higher prevalence of macromolecular infections in the zebra -fish genome.''
\end{quote}
\textbf{Analysis:} BERT appears to be hallucinating here, bringing up zebrafish as a part of the answer even though it is not a part of the question. This is the danger of Science based LLMs that may not be fully trained in the context. They can hallucinate, and it can sound convincing. 
\paragraph{SciBERT Answer}
\begin{quote}
    ``the results of this study suggest that macrolide resistance is not a common phenomenon in p. pallidum isolates.''
\end{quote}
\textbf{Analysis:} Only SciBERT directly answers the question in a straightforward, coherent way. It shows promise in using a model  with strong training in a scientific domain. 

Our models for BERT and SciBERT are also stored in Georgia Tech's HuggingFace org for reference and use. BERT is stored here \cite{georgiatech_bert_pubmedqa} and SciBERT here \cite{georgiatech_scibert_pubmedqa}

\section{Example output for NER task} \label{example_ner}
\begin{figure*}[htbp]
    \centering
    \begin{subfigure}[b]{0.45\textwidth}
        \includegraphics[width=\textwidth]{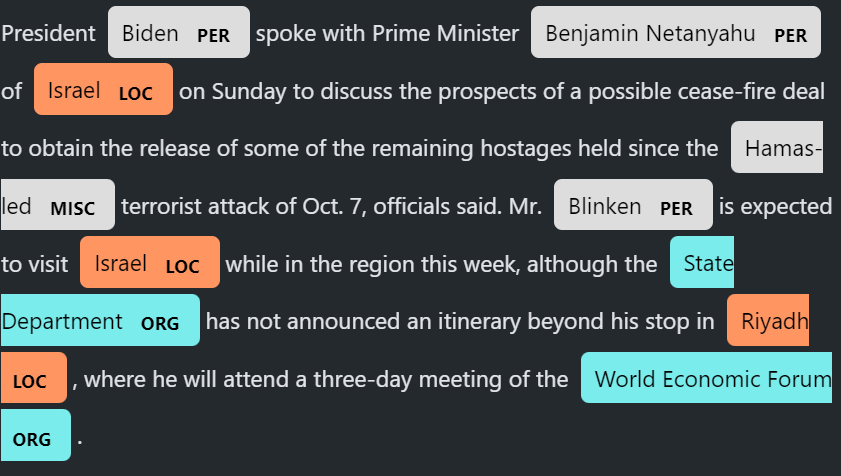}
        \caption{General}
    \end{subfigure}
    \hfill
    \begin{subfigure}[b]{0.45\textwidth}
        \includegraphics[width=\textwidth]{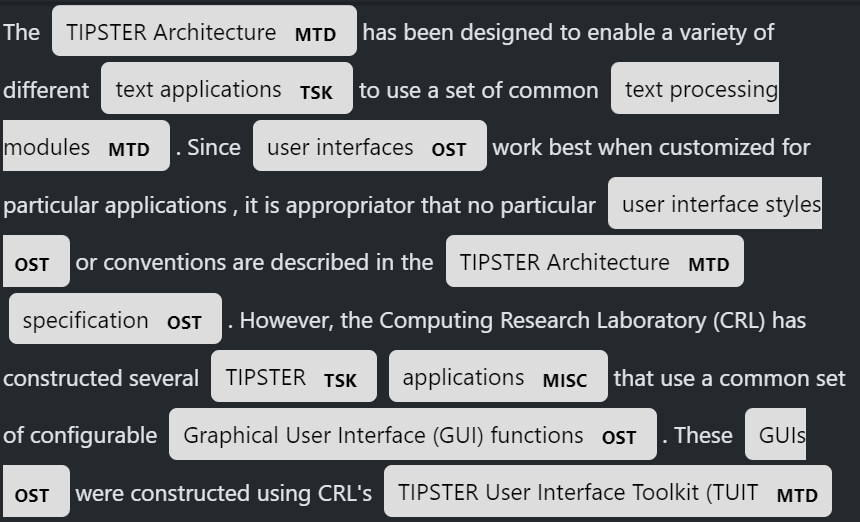}
        \caption{SciERC}
    \end{subfigure}
        \begin{subfigure}[b]{0.45\textwidth}
        \includegraphics[width=\textwidth]{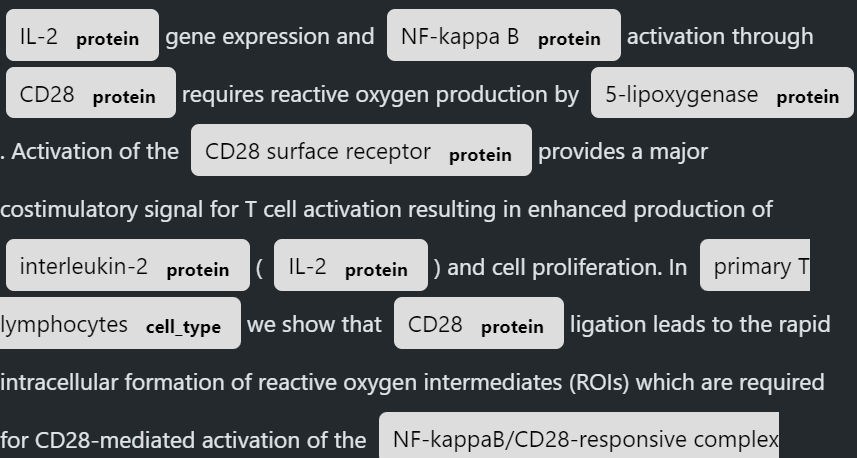}
        \caption{GENIA-bio}
    \end{subfigure}
    \caption{Example output from NER task that shows the named entities present in the input text.}
    \label{fig:ner_output}
\end{figure*}

Example output of the NER task for the three different datasets: ConLL2003, SciERC and GENIA using the best fine-tuned model based on SciDeBERTa are shown in Fig. \ref{fig:ner_output}.

\end{document}